%% file: main.tex
\documentclass[runningheads]{llncs}
\usepackage{graphicx}
\usepackage{wrapfig}
\usepackage{tikz}
\usepackage{comment}
\usepackage{amsmath,amssymb} 
\usepackage{color}
\usepackage{booktabs}
\usepackage{multirow}

\usepackage[accsupp]{axessibility}  

\usepackage[width=122mm,left=12mm,paperwidth=146mm,height=193mm,top=12mm,paperheight=217mm]{geometry}
\usepackage[pagebackref,breaklinks,colorlinks]{hyperref}
\usepackage[capitalize]{cleveref}
\crefname{section}{Sec.}{Secs.}
\Crefname{section}{Section}{Sections}
\Crefname{table}{Table}{Tables}
\crefname{table}{Tab.}{Tabs.}

\makeatletter
\newcommand{\printfnsymbol}[1]{%
  \textsuperscript{\@fnsymbol{#1}}%
}
\makeatother

\begin{document}

\pagestyle{headings}
\mainmatter

\title{AdaAfford: Learning to Adapt Manipulation Affordance for 3D Articulated Objects via Few-shot Interactions} 
\titlerunning{AdaAfford}
\author{Yian Wang\inst{1,2}\thanks{Equal contribution} \and
Ruihai Wu\inst{1,2}\printfnsymbol{1}  \and
Kaichun Mo\inst{3}\printfnsymbol{1}\\ \and
Jiaqi Ke\inst{1,2} \and
Qingnan Fan\inst{4} \and
Leonidas Guibas\inst{3}\\ \and
Hao Dong\inst{1,2,5}\printfnsymbol{4}
}

\authorrunning{Y. Wang et al.}
%
\institute{CFCS, CS Dept., PKU \and
AIIT,  PKU\\
\email{\{yianwang,wuruihai,kjq001220,hao.dong\}@pku.edu.cn}\\
\and
Stanford University\\
\email{\{kaichun,guibas\}@cs.stanford.edu}\\
\and Tencent AI Lab\\
\email{fqnchina@gmail.com}\\
\and Peng Cheng Lab\\
\url{https://hyperplane-lab.github.io/AdaAfford}
}

\maketitle



\input{tex/abs}


\input{tex/intro}

\input{tex/related}
\input{tex/problem}
\input{tex/method}

\input{tex/exps}

\input{tex/conclu}

\paragraph{Acknowledgements.}
National Natural Science Foundation of China —Youth Science Fund (No.62006006). 
Leonidas and Kaichun were supported by the Toyota Research Institute (TRI) University 2.0 program, NSF grant IIS-1763268, a Vannevar Bush Faculty Fellowship, and a gift from the Amazon Research Awards program.
The Toyota Research Institute University 2.0 program\footnote{Toyota Research Institute ("TRI") provided funds to assist the authors with their research but this article solely reflects the opinions and conclusions of its authors and not TRI or any other Toyota entity.}.

{\small
\bibliographystyle{ieee_fullname}
\bibliography{egbib}
}
\input{tex/supp_data}

\section{More Details about Method}
\vspace{-4mm}

In our setting, the subsequent interactions always continue from the previous interactions. But it actually doesn't matter for our method, we can take any distribution of test-time interactions in the training process and it won't violate our design. 

The input point cloud $O$ is the current observation and might be changed if an interaction successfully moves the object. 
\input{tex/supp_ablation}
\input{tex/supp_results}

\end{document}

%% file: tex/abs.tex
\begin{abstract}
Perceiving and interacting with 3D articulated objects, such as cabinets, doors, and faucets, pose particular challenges for future home-assistant robots performing daily tasks in human environments.

Besides parsing the articulated parts and joint parameters, researchers recently advocate learning manipulation affordance over the input shape geometry which is more task-aware and geometrically fine-grained. 
However, taking only passive observations as inputs, these methods ignore many hidden but important kinematic constraints (\emph{e.g.}, joint location and limits) and dynamic factors (\emph{e.g.}, joint friction and restitution), therefore losing significant accuracy for test cases with such uncertainties.
In this paper, we propose a novel framework, named AdaAfford, that learns to perform very few test-time interactions for quickly adapting the affordance priors to more accurate instance-specific posteriors.
We conduct large-scale experiments using the PartNet-Mobility dataset and prove that our system performs better than baselines.
{\renewcommand{\thefootnote}{\fnsymbol{footnote}} \footnotetext[4]{Corresponding author. }}

\end{abstract}

%% file: tex/intro.tex
\vspace{-4mm}
\section{Introduction}
\vspace{-2mm}
\label{sec:introduction}

For future home-assistant robots to aid humans in accomplishing diverse everyday tasks, we must equip them with strong capabilities perceiving and interacting with diverse 3D objects in human environments.
Articulated objects, such as cabinets, doors, and faucets, are particularly interesting kinds of 3D shapes in our daily lives since agents can interact with them and trigger functionally important state changes of the objects (\emph{e.g.}, push closed the drawer of the cabinet, rotate the handle and pull open the door, turn on/off the water from the faucet by rotating the switch).
However, because robots need to understand more semantically complicated part semantics and manipulate articulated parts with higher degree-of-freedoms than rigid objects, it remains a very important yet challenging task to perceive and interact with 3D articulated objects.

Many previous works have investigated the problem of perceiving and interacting with 3D articulated objects.
Researchers have been pushing the state-of-the-arts on segmenting articulated parts~\cite{tzionas2016reconstructing,lipart}, tracking them~\cite{schmidt2014dart,weng2021captra}, and estimating joint parameters~\cite{wang2019shape2motion,yan2020rpm}, enabling robotic systems~\cite{peterson2000high,chitta2010planning,urakami2019doorgym} to successfully perform sophisticated planning and control over 3D articulated objects.

\begin{figure}[ht]
    \begin{center}
        \includegraphics[width=0.7\textwidth, 
        trim={0cm, 0cm, 0cm, 0cm}, clip]{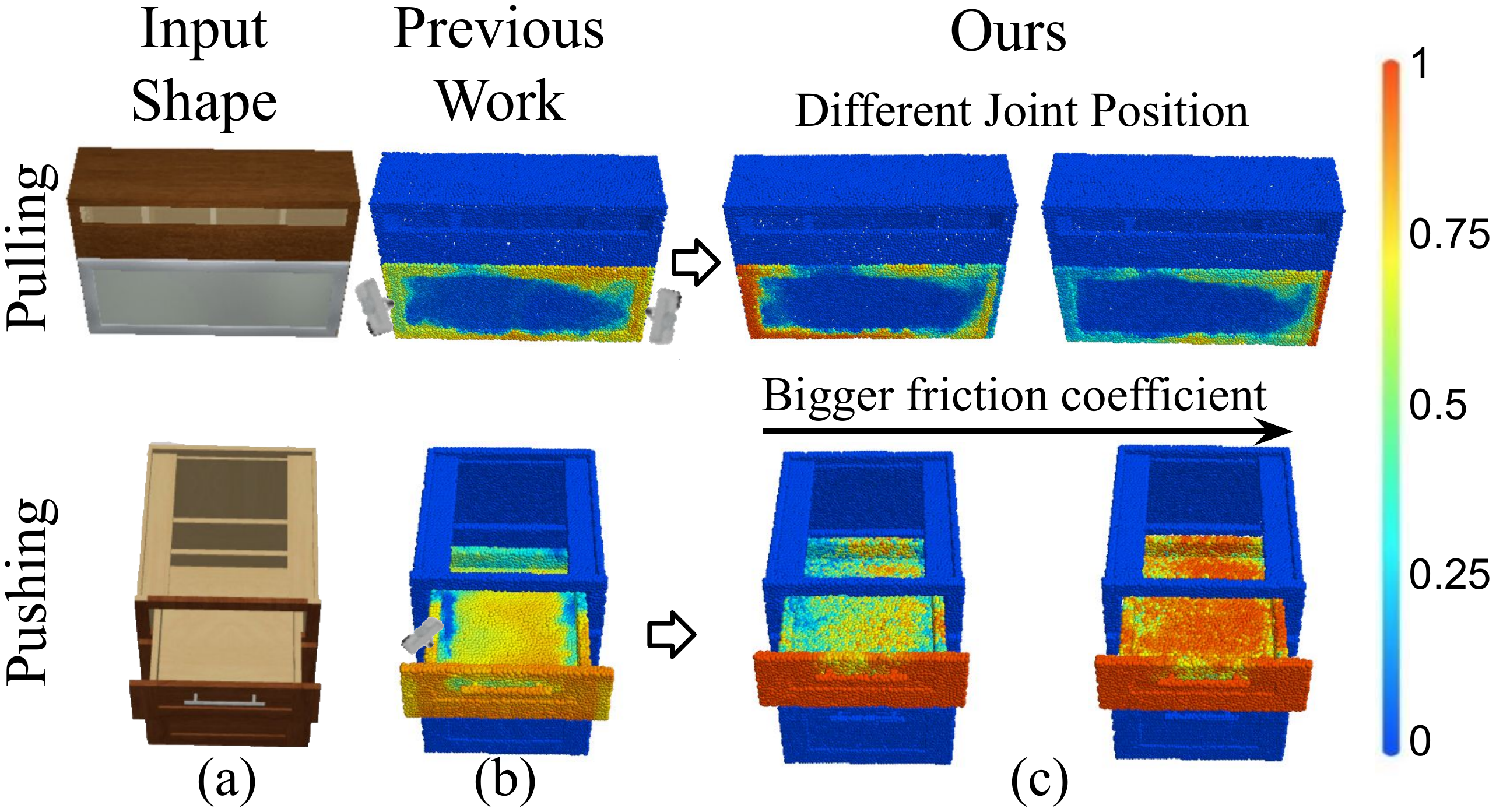}
    \end{center}
    \vspace{-3mm}
    \caption{
    For robotic manipulation over 3D articulated objects (a), past works~\cite{Mo_2021_ICCV,wu2022vatmart} have demonstrated the usefulness of per-point manipulation affordance (b).
    However, only observing static visual inputs passively, these systems suffer from intrinsic ambiguities over kinematic constraints.
    Our \textit{AdaAfford} framework reduces such uncertainties via interactions and quickly adapts instance-specific affordance posteriors (c).
    }
    \label{fig:teaser}
    \vspace{-3mm}
\end{figure}

More recently, beyond recognizing the articulated parts and joints, researchers have been proposing learning more task-aware and geometrically fine-grained manipulation affordance over input 3D geometry.
Where2Act~\cite{Mo_2021_ICCV}, the most related to our work, learns densely labeled manipulation affordance heatmaps over 3D input partial scans of articulated objects, as illustrated in Fig.~\ref{fig:teaser} (b), by performing self-supervised trial-and-error interaction in a physical simulator.
There are also many other works leveraging similar dense affordance predictions over 3D scenes~\cite{interaction-exploration} and rigid objects~\cite{mandikal2020graff}.
Such densely labeled affordance predictions over 3D data provide more geometrically fine-grained actionable information and can be learned task-specifically given different manipulation actions, showing promises in bridging the perception-interaction gaps for robotic manipulation over large-scale 3D data across different tasks.

However, taking only a single-frame observation of the 3D shape as input (\emph{e.g.}, a single 2D image, a single partial 3D scan), these methods systematically fail to capture many hidden but important kinematic or dynamic factors and therefore predict inaccurate affordance heatmaps, similar to Fig.~\ref{fig:teaser}~(b), by averaging out such uncertainties.
For example, given a fully closed cabinet door with no obvious handle as shown in Fig.~\ref{fig:teaser} (top-row), it is uncertain if the door axis is on the left or right side, which significantly affects the manipulation affordance predictions.
Other kinematic uncertainties include joint limits (\emph{e.g.}, push inward or pull outward for a door) and types (\emph{e.g.}, slide or rotate to open a door).
Besides, various dynamic or physical parameters (\emph{e.g.}, part mass, joint friction) are also unobservable from single-frame inputs but largely affect manipulation affordance.
For example, with increasing friction coefficient for a cabinet drawer (Fig.~\ref{fig:teaser}, bottom-row), robots would be able to push the inner board.

In this paper, we propose a novel framework \textit{AdaAfford} learning perform very few test-time interactions to reduce such kinematic or dynamic uncertainties and fastly adapts the affordance prior predictions to instance-specific posteriors given a novel test shape.
Our system learns a \textit{data-efficient strategy} that sequentially samples very few uncertain or interesting locations to interact, as the interacting grippers illustrated in Fig.~\ref{fig:teaser}~(b), according to the current affordance predictions and past interaction trials (we begin with the affordance prior predictions of Where2Act~\cite{Mo_2021_ICCV} and zero interaction history).
The interaction outcomes, each of which includes the interaction location, direction, and the resulting part motion, are then observed and incorporated to produce posterior affordance predictions, as illustrated in Fig.~\ref{fig:teaser}~(c), by a proposed \textit{fast-adaptation mechanism}.
We set up a benchmark for experiments and evaluations using the large-scale PartNet-Mobility dataset~\cite{Mo_2019_CVPR} and the SAPIEN physical simulator~\cite{Xiang_2020_SAPIEN}.
We use in total 972 shapes from 15 object categories and conduct experiments for several action types, 
and randomly sample the kinematic and dynamic parameters for the 3D articulated objects in simulation. 
Experiments show our method can successfully and efficiently adapt manipulation affordance to novel test shapes with as few as one to four interactions.
Quantitative evaluation  
further proves the effectiveness of our proposed approach.

In summary, our main contributions are the following.
    1) we point out and investigate an important limitation of the methods that learn densely labeled visual manipulation affordance -- the unawareness of hidden yet important kinematic and dynamic uncertainties;
    2) we propose a novel framework \textit{AdaAfford} that learns to perform very few test-time interactions to reduce uncertainties and quickly adapt to predicting an instance-specific affordance posterior;
    3) we set up a large-scale benchmark, built upon PartNet-Mobility~\cite{Mo_2019_CVPR} and SAPIEN~\cite{Xiang_2020_SAPIEN}, for experiments and evaluations, and results demonstrated the effectiveness and efficiency of the proposed approach.

%% file: tex/related.tex
\vspace{-2mm}
\section{Related Work}
\vspace{-2mm}
\label{sec:related_work}

\vspace{-1mm}
\paragraph{Visual Affordance on 3D Shapes.}
Affordance~\cite{gibson1977theory} suggests possible ways for agents to interact with objects.
Many past works have investigated learning 
grasp~\cite{redmon2015real,lenz2015deep,qin2020s4g,kokic2020learning,jiang2021synergies} and 
manipulation~\cite{interaction-exploration,qin2020keto,mandikal2020graff,Mo_2021_ICCV,wu2022vatmart,xu2022umpnet} 
affordance for robot-object interaction, while there are also many works studying affordance for
hand-object~\cite{kjellstrom2011visual,fang2018demo2vec,mandikal2020graff,yang2021cpf,corona2020ganhand},
object-object~\cite{sun2014object,zhu2015understanding,mo2021o2oafford}, 
and human-scene~\cite{fouhey2012people,3d-affordance,qi2020learning,interaction-exploration} interaction scenarios.
Among these works, researchers have proposed different representations for visual affordance, including detection locations~\cite{redmon2015real,lenz2015deep}, parts~\cite{mandikal2020graff}, keypoints~\cite{qin2020keto}, heatmaps~\cite{interaction-exploration,Mo_2021_ICCV}, etc.
In this work, we mostly follow the settings in~\cite{Mo_2021_ICCV} for learning visual affordance heatmaps for manipulating 3D articulated objects.
Different from previous works that infer possible agent-object visual affordance heatmaps passively from static visual observations, our framework leverages active interactions to efficiently query uncertain kinematic or dynamic factors for learning more accurate instance-adaptive visual affordance.

\vspace{-2mm}
\paragraph{Fast Adaption via Few-shot Interactions.}
Researchers have explored various approaches~\cite{zhou19epi,finn2017model,rakelly2019efficient,zhao2020meld,farid2021few} for fast adaption via few-shot interactions.
Many past works have also designed interactive perception methods to figure out object mass~\cite{kumar2019estimating}, dynamic parameters~\cite{xu2019densephysnet,agrawal2016learning,ferreira2019learning,janner2018reasoning}, or parameters for known models~\cite{yu2017preparing}.
Different from these studies proposing general algorithms for policy adaptation or figuring out explicit system parameters for rigid objects, we focus on designing a working solution for our specific task of learning visual affordance heatmaps for manipulating 3D articulated objects with special designs on predicting geometry-grounded interaction proposals and interaction-adaptive affordance predictions.





%% file: tex/problem.tex
\vspace{-2mm}
\section{Problem Formulation}
\vspace{-2mm}
\label{sec:problem}

Given as input a single-frame 3D partial point cloud observation of an articulated object $O\in\mathbb{R}^{N\times3}$ (\emph{e.g.}, lifted from a depth scanner with known camera intrinsics), the Where2Act framework~\cite{Mo_2021_ICCV} directly outputs a per-point manipulation affordance heatmap $A\in[0,1]^{N}$, where higher scores indicate bigger chances for being interacted with to accomplish a given short-term manipulation task (\emph{e.g.}, pushing, pulling).
Additionally, a diverse set of gripper orientations $\{R_1^p,R_2^p,\cdots|R_i^p\in SO(3)\}$ is proposed at each point $p\in O$ suggesting possible ways for robot agents to interact with, each of which also associated with a success likelihood $s_i^p\in[0,1]$.
No interaction is allowed at test time in Where2Act and a fixed set of system dynamic parameters is used across all shapes.

We follow most of the Where2Act settings except that we randomly vary the system dynamics and allow test-time interactions over the 3D shape to reduce kinematic or dynamic uncertainties.
Our AdaAfford system proposes a few interactions sequentially $\mathcal{I}=\{I_1,I_2,\cdots\}$.
Each interaction $I_i=(O_i, p_i, R_i, m_i)$ executes a task-specific hard-coded short-term trajectory defined in Where2Act, parametrized by the interaction point $p_i\in O_i$ and the gripper orientation $R_i\in SO(3)$, and observes a part motion $m_i$.
Starting from the input shape observation $O_1\leftarrow O$, every interaction $I_i$ where $m_i\ne0$ changes the part state and thus produces a new shape point cloud input for the next interaction $O_{i+1}\ne O_i$.
Leveraging the interaction observations $\mathcal{I}$, our system then adapts the per-point manipulation affordance $A$ predicted by Where2Act to a posterior $A_{\mathcal{I}}\in[0,1]^{N}$ that reduces uncertainties and provides more accurate instance-specific predictions.
For each gripper orientation $R_i$, we also update the success likelihood score $s_{i,\mathcal{I}}^p\in[0,1]$ considering the test-time interactions.

%% file: tex/method.tex
\vspace{-2mm}
\section{Method}
\vspace{-2mm}
\label{sec:method}

Our proposed \textit{AdaAfford} framework primarily consists of two modules -- an \textit{Adaptive Interaction Proposal} (AIP) module and an \textit{Adaptive Affordance Prediction} (AAP) module.
While the AIP module learns a greedy yet effective strategy for sequentially proposing few-shot test-time interactions $\mathcal{I}=\{I_1,I_2,\cdots\}$, the AAP module is trained to adapt affordance predictions from Where2Act~\cite{Mo_2021_ICCV} prior $A$ to a posterior $A_{\mathcal{I}}$ observing the sampled interactions $\mathcal{I}$.
We iterate two modules recurrently at test time to produce a sequence of few-shot interactions $\mathcal{I}$ leading to the final affordance posterior prediction $A_\mathcal{I}$.
During training, we iteratively alternate the training for the two modules until a joint convergence.


Below, we first introduce the test-time inference procedure for a brief overview.
Next, we describe the input backbone encoders that are shared among all networks in our framework.
Then, we describe the detailed architectures and system designs of the two modules.
We conclude with the training losses and strategy.


\begin{figure}[t]
    \begin{center}
    \includegraphics[width=0.75\textwidth, 
        trim={0cm, 0.1cm, 0cm, 0.1cm}, clip]{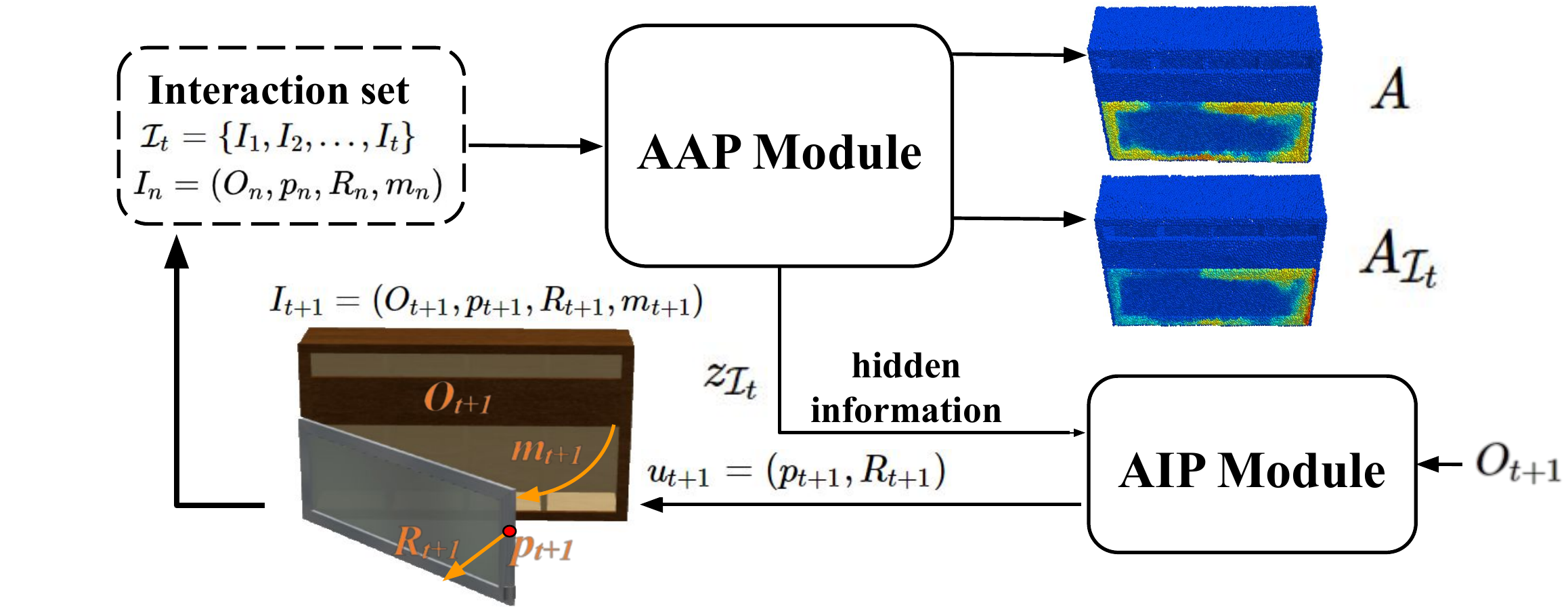}
    \end{center}
    \vspace{-3mm}
    \caption{\textbf{Method Overview.} 
    Starting from the Where2Act~\cite{Mo_2021_ICCV} predicted affordance prior $A$, at each timestep $t=1,2,\cdots$, we recursively leverage the \textit{Adaptive Interaction Proposal} (AIP) module to propose a next-time interaction action $u_{t+1}$ 
    , observe the interaction outcome $m_{t+1}$, and feed through the \textit{Adaptative Affordance Prediction} (AAP) module all past few-shot interactions $\mathcal{I}_t$ together with the new one $I_{t+1}$ for adapting to an affordance posterior prediction $A_{\mathcal{I}_{t+1}}$.
    The procedure iterates until the interaction budget is reached or the AIP module decides to stop.
    }
    \vspace{-3mm}
    \label{fig:method_overview}
\end{figure}

\vspace{-2mm}
\paragraph{Test-time Overview.}
Fig.~\ref{fig:method_overview} presents an overview of the method.
We apply a recurrent structure at test time. 
Starting from the affordance prediction $A$ without any interaction, the AIP module proposes the first action for producing the interaction data $I_1$.
Then, at each timestep $t=1,2,\cdots$, we feed the current set of interactions $\mathcal{I}_t=\{I_{1},\cdots, I_{t}\}$ as inputs to the AAP module and extract hidden information $z_{\mathcal{I}_t}\in\mathbb{R}^{128}$ that adapts the affordance map prediction to $A_{\mathcal{I}_t}$. 
The AIP module then takes $z_{\mathcal{I}_t}$ as input and proposes an action $u_{t+1}=(p_{t+1},R_{t+1})$ composed of the interaction point $p_{t+1}$ and the gripper orientation $R_{t+1}$ for the next interaction. 
Performing this action in the environment, we obtain the next-step interaction data $I_{t+1}=(O_{t+1}, p_{t+1}, R_{t+1}, m_{t+1})$ and put it into the interaction set $\mathcal{I}_{t+1}\leftarrow\mathcal{I}_t\cup\{I_{t+1}\}$.
We iterate until the interaction budget has been reached or our AIP module decides to stop.
When the procedure stops at timestep $T$, we output the final affordance posterior $A_\mathcal{I}=A_{\mathcal{I}_T}$.

\vspace{-2mm}
\paragraph{Input Encoders.}
This paragraph details how we encode inputs into features as all the encoder networks in the two modules take the same input entities (\emph{e.g.}, the shape observation $O$, the interaction action $u$) and thus share the same architecture. 
We use the PointNet++ segmentation network~\cite{qi2017pointnetplusplus} to encode the input shape point cloud $O\in\mathbb{R}^{N\times3}$ into per-point feature maps $f_O\in\mathbb{R}^{N\times 128}$ and denote $f_{p|O}\in \mathbb{R}^{128}$ as the feature at any point $p\in O$. 
We use Multilayer Perceptron (MLP) networks to encode other vector inputs (\emph{e.g.}, the interaction action $u$ and the part motion $m$) into $f_{a}\in R^{128}$.
The networks in the following subsections will first encode the inputs into $f_{p|O}$ and $f_{a}$, and then concatenate them into $f_{I} \in R^{256}$.
The encoders do not share weights across different modules.



\begin{figure*}[t]
    \begin{center}
        \includegraphics[width=\linewidth]{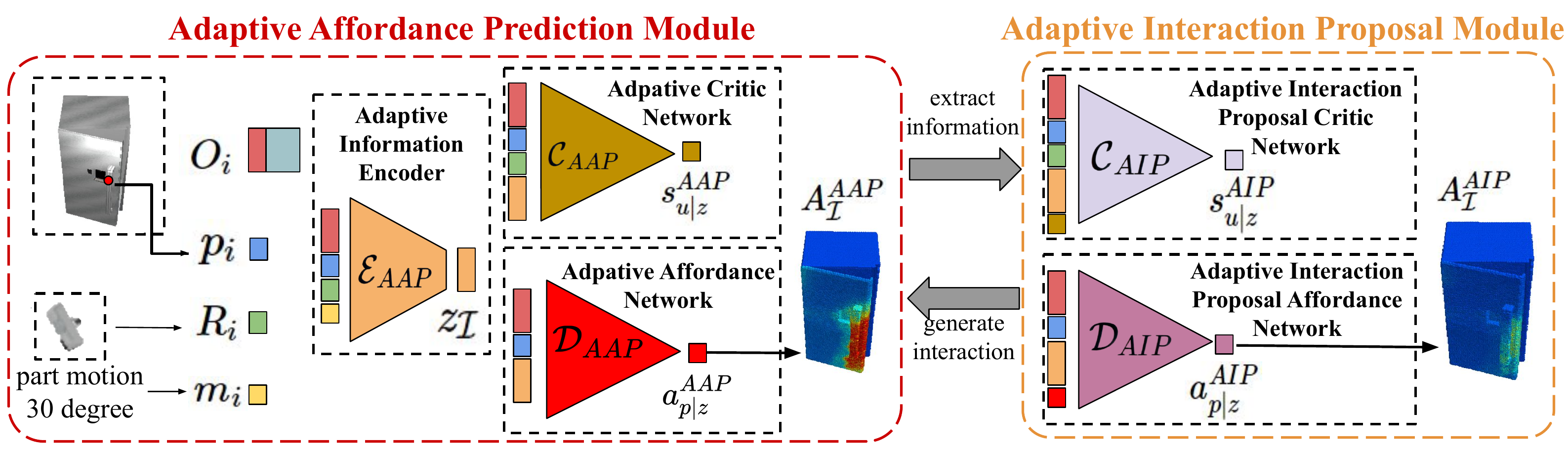}
    \end{center}
    \vspace{-3mm}
    \caption{\textbf{Network Architecure.}
    Left: the \textit{Adaptive Affordace Prediction} (AAP) module takes as inputs the few-shot interactions $\mathcal{I}$ and predicts the affordance posterior $A_\mathcal{I}$.
    Right: the \textit{Adapative Interaction Proposal} (AIP) module proposes a next-step interaction action $u_{t+1}=(p_{t+1}, R_{t+1})$ (denote the current timestep as $t$) given the feature $z_\mathcal{I}$ extracted from the current interaction observations $\mathcal{I}$.
    }
    \vspace{-3mm}
    \label{fig:train}
\end{figure*}

\vspace{-2mm}
\subsection{Adaptive Affordance Prediction Module}
\vspace{-1mm}
The \textit{Adaptive Affordance Prediction} (AAP) module takes as inputs few-shot interactions $\mathcal{I}$ and predicts the affordance posterior $A_\mathcal{I}$.
This module is composed of three subnetworks: 1) an \textit{Adaptive Information Encoder} $\mathcal{E}_{AAP}$ that extracts hidden information $z\in \mathbb{R}^{128}$ from a set of interactions $\mathcal{I}$; 2) an \textit{Adaptive Affordance Network} $\mathcal{D}_{AAP}$ that predicts the posterior affordance heatmap $A_\mathcal{I}$ conditioned on the hidden information $z$; and 3) an \textit{Adaptive Critic Network} $\mathcal{C}_{AAP}$ that predicts the AAP action score $s_{u|z}^{AAP}\in[0,1]$ for an action $u$ conditioned on the hidden information $z$. Here, an action is represented as $u=(p,R)$ including an interaction point $p\in O$ and a gripper orientation $R\in SO(3)$.

\vspace{-2mm}
\paragraph{Adaptive Information Encoder.} Given a set of interactions $\mathcal{I}=\{I_1,I_2,\cdots\}$ as inputs, the \textit{Adaptive Information Encoder} $\mathcal{E}_{AAP}$ outputs a 128-dim hidden information representation $z_\mathcal{I}$ ($z$ for brevity). It first encodes each interaction $I_{i}$ using the input encoders mentioned before, and then uses an MLP network to encode the features into a 128-dim latent code $z_{I_{i}}$ representing the hidden information extracted from $I_{i}$. As different interactions contain different amount of hidden information, we use another MLP Network to predict an attention score $w_{I_i}\in\mathbb{R}$ for each interaction. To get a summarized hidden information from a set of interactions, we simply computes a weighted average over all $z_{I_{i}}$'s according to the weights $w_{I_i}$'s and use the resulting feature as $z_{\mathcal{I}}$. Formally, we have $z_{\mathcal{I}}\leftarrow\left(\sum_i z_{I_{i}} \times w_{I_i}\right) / \left(\sum_i w_{I_i}\right)$.

\vspace{-2mm}
\paragraph{Adaptive Critic Network.} Given 
the object partial point cloud observation $O$, an arbitrary interaction point $p\in O$, an arbitrary gripper orientation $R\in SO(3)$ and the latent code $z$, 
the \textit{Adaptive Critic Network} $\mathcal{C}_{AAP}$ predicts an AAP action score $s_{u|z}^{AAP}\in[0,1]$ indicating the likelihood for the success of the interaction action $u$ given the interaction information $z$. 
It first encodes the input $\{O,P,R\}$ using the input encoders as mentioned before and then employs an MLP network to predict AAP action score $s_{u|z}^{AAP}$, taking the concatenated features together with $z$ as inputs.
A higher AAP action score $s_{u|z}^{AAP}$ for action $u$ indicates a higher chance for $u$ to succeed in accomplishing the given manipulation task.

\vspace{-2mm}
\paragraph{Adaptive Affordance Network.} Given the input object partial point cloud $O$, an arbitrary point $p\in O$, and the latent code $z$, the \textit{Adaptive Affordance Network} $\mathcal{D}_{AAP}$ predicts an actionability score $a_{p|z}^{AAP}\in[0,1]$ at point $p$.
It first encodes the input $\{O,p\}$ using the aforementioned input encoders and then uses an MLP network that takes the concatenated features together with $z$ as inputs and produces an actionability score $a_{p|z}^{AAP}$ as the output.
A higher actionability score $a_{p|z}^{AAP}$ indicates a higher chance to successfully interact on point $p$.

\vspace{-2mm}
\subsection{Adaptive Interaction Proposal Module}
\vspace{-1mm}
\textit{Adaptive Interaction Proposal} (AIP) module proposes an action (denote the current timestep as $t$) $u_{t+1}=(p_{t+1}, R_{t+1})$ for the next step interaction, given the feature $z$ extracted from the current interaction observations $\mathcal{I}$.
This module contains two networks: 1) an \textit{Adaptive Interaction Proposal Affordance Network} $\mathcal{D}_{AIP}$ that predicts an AIP actionability score $a_{p|z}^{AIP}\in\mathbb{R}$ indicating how likely the next-action is worth interacting at point $p$, and 2) an \textit{Adaptive Interaction Proposal Critic Network} $\mathcal{C}_{AIP}$ predicting an AIP action score $s_{u|z}^{AIP}\in\mathbb{R}$ suggesting the gripper orientation to pick for the next interaction.
We leverage the predictions of the two networks to propose the next action $u_{t+1}=(p_{t+1}, R_{t+1})$.



\vspace{-2mm}
\paragraph{Adaptive Interaction Proposal Critic Network.} 
Given the input object partial point cloud $O$, an arbitrary interaction point $p\in O$, an arbitrary gripper orientation $R\in SO(3)$, the latent code $z$, and the AAP action score $s_{u|z}^{AAP}$ produced by $\mathcal{C}_{AAP}$, the \textit{AIP Critic Network} $\mathcal{C}_{AIP}$ predicts the AIP action score $s_{u|z}^{AIP}\in\mathbb{R}$ of $u$.
It first encodes the inputs $\{O,p,R,s_{u|z}^{AAP}\}$ using the input encoders and then uses an MLP network that takes the concatenated features together with $z$ as inputs and generates an AIP action score $s_{u|z}^{AIP}$ for the action $u$.
A higher AIP action score suggests that the action $u$ may query more unknown yet interesting hidden information and thus is worth exploring next.

\vspace{-2mm}
\paragraph{Adaptive Interaction Proposal Affordance Network.} 
Given the input partial shape observation $O$, an arbitrary interaction point $p\in O$, the latent code $z$, and the AAP actionability score $a_{p|z}^{AAP}$ at point $p$ estimated by $\mathcal{D}_{AAP}$, the \textit{AIP Affordance Network} $\mathcal{D}_{AIP}$ predicts the AIP actionability score $a_{p|z}^{AIP}\in\mathbb{R}$ at point $p$.
It first encodes the inputs $\{O,p,a_{p|z}^{AAP}\}$ using the aforementioned input encoders and then employs an MLP network to predict an AIP actionability score $a_{p|z}^{AIP}$, taking the concatenated features together with $z$ as inputs.
A higher AIP actionability score at $p$ indicates more unknown yet helpful hidden information may be obtained by executing an interaction at $p$. 

\vspace{-2mm}
\paragraph{Next-step Interaction Action Proposal.}
To propose an action $u_{t+1}=(p_{t+1}, R_{t+1})$ for the next interaction, given the hidden information $z$ and the input shape partial point cloud $O$, we first obtain the AIP actionability heatmap $A_{p|z}^{AIP}$ for every point $p\in O$ predicted by the \textit{AIP Affordance Network} $\mathcal{D}_{AIP}$ and then select the point $p_{t+1}\leftarrow p_*$ with the highest AIP actionability score $a_{p_*|z}^{AIP}$. Then, we sample 100 random actions $\{u_1,u_2,\cdots,u_{100}\}$ at $p$ using the Where2Act's pre-trained \textit{Action Proposal Network}, use our \textit{AIP critic network} $\mathcal{C}_{AIP}$ to generate the AIP action scores $s_{u_i|z}^{AIP}$ for each action $u_i$, and then choose the action $u_{t+1}\leftarrow u_*$ with the highest AIP action score $s_{u_*|z}^{AIP}$.

\vspace{-2mm}
\paragraph{Stopping Criterion for the Few-shot Interactions.}
The AIP procedure for generating few-shot interactions stops when a preset budget is reached or the maximal AIP actionability score is below a certain threshold (\emph{e.g.}, 0.05).

\vspace{-3mm}
\subsection{Training and Losses}
\vspace{-3mm}

In brief, for AAP module, we use ground-truth motion $m$ to supervise $\mathcal{E}_{AAP}$ and $\mathcal{C}_{AAP}$, and utilize $\mathcal{C}_{AAP}$ to supervise the training of $\mathcal{D}_{AAP}$. For AIP module, we use AAP module to supervise the training of $\mathcal{C}_{AIP}$ and use it to supervise $\mathcal{D}_{AIP}$.
Below, we describe the losses and the training strategy in detail.

\vspace{-2mm}
\paragraph{AAP Action Scoring Loss.} 
To supervise $\mathcal{C}_{AAP}$, we use a standard binary cross entropy loss, which measures the error between the prediction of $\mathcal{C}_{AAP}$ and target part's ground truth motion $m$ of an interaction $I$. Specifically, given the hidden information $z$, a batch of interaction observations $\mathcal{I}=\{I_1,I_2,...,I_B\}$ where $I_i=\{O_i,u_i,m_i\}$, and the AAP action score prediction $s_{u_i|z}^{AAP}$ for each interaction $I_i$, the loss is defined as

\begin{equation}
\mathcal{L}_{\mathcal{C}}^{AAP}=-\frac{1}{B}\sum_i{r_i\log(s_{u_i|z}^{AAP}) + (1-r_i)\log(1-s_{u_i|z}^{AAP})}\nonumber
\end{equation}
where $r_i=1$ if $m_i>\tau$ (\emph{e.g.}, $\tau=0.01$) or $r_i=0$ rendering a binary discretization for each interaction outcome.

\vspace{-2mm}
\paragraph{AAP Actionability Scoring Loss.} 
To train the \textit{Adaptive Affordance Network} $\mathcal{D}_{AAP}$, we apply an $\mathcal{L}_{1}$ loss to measure the difference from the predicted score $a_{p|z}^{AAP}$ to the ground truth. To estimate the ground truth actionability score for $p$, we randomly sample 100 actions at $p$ according to the pre-trained Where2Act \textit{Action Proposal Network}, predict the AAP action scores $s_{u|z}^{AAP}$'s of these actions $u$'s using the \textit{Adaptive Critic Network} $\mathcal{C}_{APP}$, and take the average of the top-5 scores as the ground truth actionability score.

\vspace{-2mm}
\paragraph{AIP Action Scoring Loss.} 
To supervise the \textit{AIP Critic Network} $\mathcal{C}_{AIP}$, we use an $\mathcal{L}_{1}$ loss to measure the difference between our predicted AIP action score $s_{u|z}^{AIP}$ and the ground truth AIP action score $gt_{u|z}^{AIP}$.
We design a greedy yet effective way to estimate the ground-truth scores.
Given a set of interactions $\mathcal{I}_{T}=\{I_1,I_2,\cdots\}$, to generate $gt_{u_{i}|z}^{AIP}$ for an interaction action $u_{i}$, we respectively encode two interaction subsets $\mathcal{I}_{i-1}=\{I_1,I_2,\cdots,I_{i-1}\}$ and $\mathcal{I}_{i}=\{I_1,I_2,\cdots,I_{i}\}$ into latent codes $z_{\mathcal{I}_{i}}$ and $z_{\mathcal{I}_{i-1}}$. Then, we feed $z_{\mathcal{I}_{i}}$ and $z_{\mathcal{I}_{i-1}}$ as the conditional inputs to the \textit{Adaptive Critic Network} $\mathcal{C}_{AAP}$ separately and count the difference of the AAP action scoring loss $\mathcal{L}_{\mathcal{C}}^{AAP}$ as the ground truth of AIP action score $gt_{u_{i}|z_{\mathcal{I}_{i-1}}}^{AIP}$. More concretely, let the AAP action scoring loss conditioned on $z_{\mathcal{I}_{i}}$ and $z_{\mathcal{I}_{i-1}}$ respectively be $\mathcal{L}_{\mathcal{I}_{i}}$ and $\mathcal{L}_{\mathcal{I}_{i-1}}$. We define the ground truth AIP action score $gt_{u_{i}|z_{\mathcal{I}_{i-1}}}^{AIP}\leftarrow\mathcal{L}_{\mathcal{I}_{i-1}} - \mathcal{L}_{\mathcal{I}_{i}}$.
The AIP action score is trained to regress an estimated positive influence of executing $u$ on the AAP action score predictions, where an action giving more influence is preferred as it helps discover more hidden information useful to the task.


\vspace{-2mm}
\paragraph{AIP Actionability Scoring Loss.} 
To train the \textit{AIP Affordance Network} $\mathcal{D}_{AIP}$, we use another $\mathcal{L}_{1}$ loss.
For each position $p\in O$,
we sample 100 actions $u_i$'s using the pre-trained Where2Act \textit{Action Proposal Network}, obtain the AIP action scores $s_{u_i|z}^{AIP}$'s of these actions $u_i$'s predicted by the \textit{AIP Critic Network} $\mathcal{C}_{AIP}$, and use the average of the top-5 scores as the regression target.


\vspace{-2mm}
\paragraph{Training Strategy.}
We iteratively train the AAP module and AIP module until a joint convergence since the update of the subnetworks in one module will affect the training of the subnetworks in the other module.
More specifically, the update of $\mathcal{C}_{AAP}$ and $\mathcal{D}_{AAP}$ in the AAP module will affect the ground-truth AIP action scores, while the update of $\mathcal{C}_{AIP}$ and $\mathcal{D}_{AIP}$ in the AIP module will change the proposed interactions used to generate $z$ in the AAP module.
Therefore, our final solution is to train the AAP and AIP modules iteratively. 
The procedure starts with firstly training the AAP module using randomly sampled interactions.
We then train the AIP module to learn to propose more efficient and effective proposals.
Then, with the trained subnetworks in the AIP module, we finetune the AAP module with the proposed few-shot interactions.
We iteratively alternate the training until both modules converge.

%% file: tex/exps.tex
\vspace{-4mm}
\section{Experiments}
\vspace{-2mm}
\label{sec:exps}

We perform experiments using the large-scale PartNet-Mobility dataset~\cite{Mo_2019_CVPR} and the SAPIEN 
simulator~\cite{Xiang_2020_SAPIEN}, 
and set up several baselines for comparisons.
Results demonstrate the effectiveness and superiority of the proposed approach.

\vspace{-4mm}
\subsection{Data and Settings}

\vspace{-1mm}
\paragraph{Data.} 
Following the settings of Where2Act~\cite{Mo_2021_ICCV}, we conduct our experiments in the SAPIEN~\cite{Xiang_2020_SAPIEN} simulator equipped with NVIDIA PhysX~\cite{physx} simulation engine and the large scale PartNet-Mobility~\cite{Mo_2019_CVPR} dataset. We use 972 articulated 3D objects covering 15 object categories, mostly following Where2Act, to carry out the experiments. The dataset is divided into 10 training and 5 testing categories.
The shapes in the training categories are further divided into two disjoint sets of training and test shapes.
See supplementary for detailed statistics.

\vspace{-2mm}
\paragraph{Experiment Settings.} 
Following Where2Act~\cite{Mo_2021_ICCV}, we perform experiments over all object categories under different manipulation action types.
We train one network for each downstream manipulation task over training shapes from the 10 training object categories and evaluate the performance over test shapes from the training categories and shapes from unseen test categories.
Besides, to further demonstrate the effectiveness of our method, we conduct two additional experiments under challenging tasks with clear kinematic ambiguity, each of which is conducted over a single object category: 1) pulling closed doors of cabinets that cannot be easily distinguished which side to pull open; 2) pushing faucets with uncertainties which direction to rotate (clockwise or counter-clockwise).
These 
experiments are particularly interesting yet challenging cases on which previous work Where2Act~\cite{Mo_2021_ICCV} 
fail drastically and we hope to test our framework.


\input{tabs/twotable}

\vspace{-2mm}
\paragraph{Environment Settings.}
Following Where2Act, we abstract away the robot arm and only use a Franka Panda flying gripper as the robot actuator.
The input shape point cloud is assumed to be cleanly segmented out.
To generate the input partial point cloud scans, we mount an RGB-D camera with known intrinsic parameters 5-unit-length away pointing to the center of the target object.

To simulate manipulating shapes with uncertain dynamics, we randomly change the following three physical parameters in SAPIEN: 1) the friction of the target part joint, 2) the mass of the target part, and 3) the friction coefficient of the target part surface.
For the "pulling closed door" task, we manually select the cabinets whose doors have no clear handle geometry in the PartNet-Mobility dataset~\cite{Xiang_2020_SAPIEN}, and set the poses of those doors to be closed. The gripper cannot tell which side to pull open the door because it is impossible to tell whether the axis position is on the left or right of the door from passive visual observations.
For the "pushing faucet" task, we randomly set the rotating direction of the faucet switch to be in one of the following three modes: only clockwise, only counter clockwise, or both ways.

\vspace{-3mm}
\subsection{Baselines and Evaluation Metrics.} 
\vspace{-1mm}
We set up several baseline and employ two metrics for quantitative comparisons.

\vspace{-2mm}
\paragraph{Baselines and Ablation Study.}
We compare our framework with several baselines (see supplementary for more detailed descriptions for the baseline designs): 
\vspace{-1mm}
\begin{itemize}
\item[-]
\textbf{Where2Act:} the original method proposed in~\cite{Mo_2021_ICCV} where only the pure visual information is used for predicting the visual actionable information and no interaction data is used at all during test time;
\item[-]
\textbf{Where2Act-interaction:} the Where2Act method augmented with four additional interaction observations as inputs where the interaction positions are uniformly sampled over the predicted affordance heatmap using Furthest Point Sampling (FPS) and we train an additional encoding branch similar to the \textit{Adaptive Information Encoder} to extract the additional input feature;

\item[-]
\textbf{Where2Act-adaptation:} the Where2Act method augmented with a heuristic based adaptation mechanism to replace the AAP module where given the interaction observations we locally adjust the predictions for similar points;
\item[-]
\textbf{Ours-random:} a variant of our proposed method that we use randomly sampled interaction trials over the geometry instead of the AIP proposals;
\item[-]
\textbf{Ours-fps:} a variant of our proposed method that we use FPS to sample over the predicted affordance for interactions instead of the AIP proposals.
\end{itemize}
We compare to \textbf{Where2Act} to show that the few-shot interactions indeed help to remove ambiguities and improve the performance.
The \textbf{Where2Act-interaction} baseline uses FPS to sample interaction positions relying on the intuition that it may sparsely sample over all possible regions of uncertainties.
Comparing to this baseline helps validate that our iterative framework learns smarter strategies to perform more effective interaction trials.
Furthermore, the \textbf{Where2Act-adaptation} baseline helps substantiate the effectiveness of our proposed AAP module, while the \textbf{Ours-random} and \textbf{Ours-fps} baselines are designed to verify the usefulness of the proposed AIP module.

Besides, we compare to an ablated version of our method to verify the significance of iterative training between the AAP module and the AIP module.
\vspace{-4mm}
\begin{itemize}
\item[-]
\textbf{Ours w/o iter:} an ablated version that trains the whole framework without the iteratively training process.
\end{itemize}

\begin{figure}[t]
    \begin{center}
        \includegraphics[width=\linewidth]{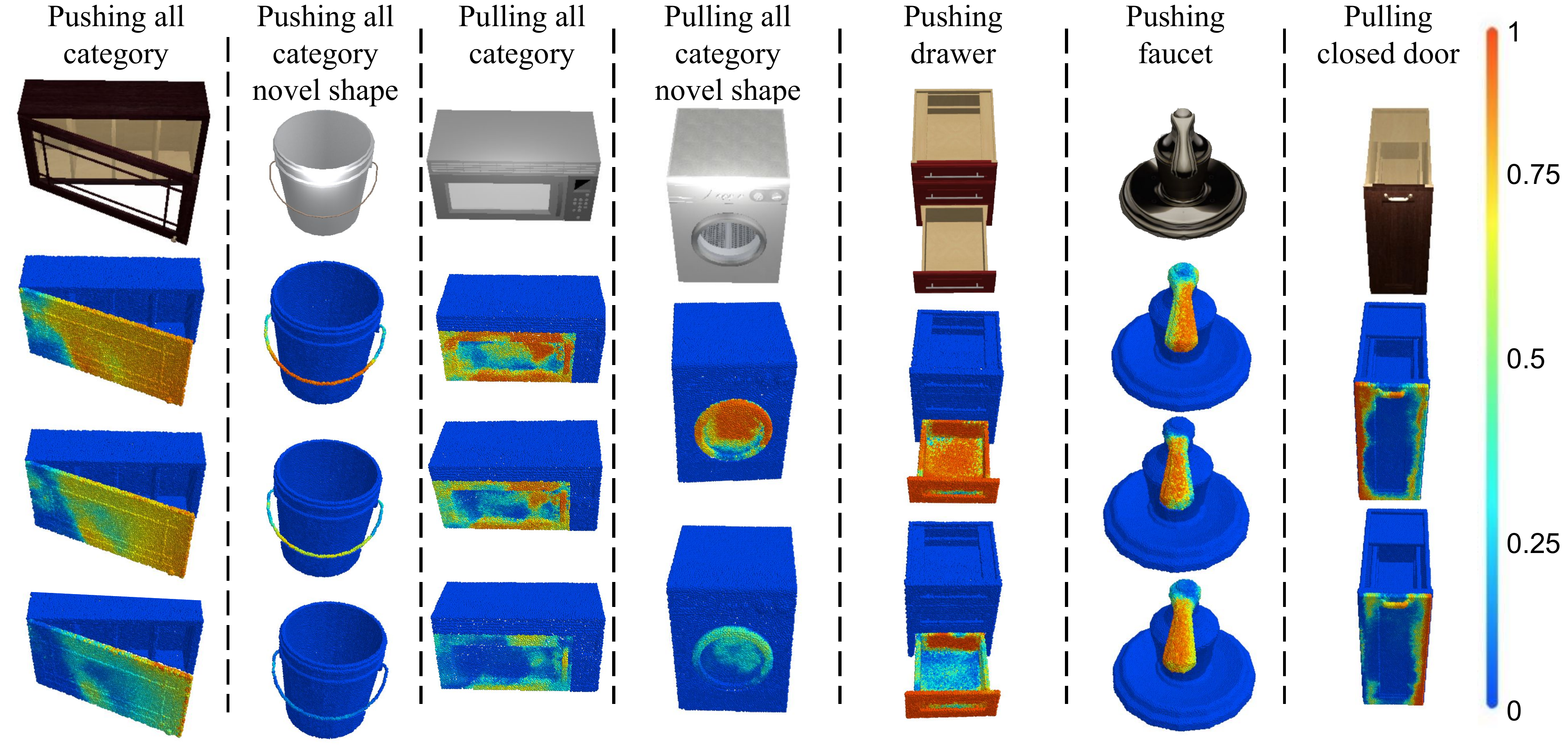}
    \end{center}
    \vspace{-3mm}
    \caption{
    Example results of adapted affordance predictions given by AAP module under different kinematic and dynamic parameters. The first five columns show the adapted affordance prediction conditioned on increasing joint friction (the first and second columns), part mass (the third column), and friction coefficient on object surface (the fourth and fifth columns). The last two columns respectively show the influence of different rotating directions (\emph{i.e.,} joint limits) and joint axis locations.
    }
    \vspace{-3mm}
    \label{fig:aff}
\end{figure}


\vspace{-2mm}
\paragraph{Evaluation Metrics.}
Following Where2Act~\cite{Mo_2021_ICCV}, we use the F-score, balancing the precision and recall, to evaluate the predictions of the \textit{Adaptive Critic Network} $\mathcal{C}_{AAP}$, and use the sample-successful rate (Sample-Succ) to evaluate the performance of the  \textit{Adaptive Critic Network} $\mathcal{C}_{AAP}$ and the \textit{Adaptive Affordance Network} $\mathcal{D}_{AAP}$. 
To compute the sample successful rate, we apply the learned test-time strategy to fill $\mathcal{I}$ and then use the extracted hidden information $z$ as the conditional input to $\mathcal{C}_{AAP}$ and $\mathcal{D}_{AAP}$. After that, we randomly select a point to interact from the group of points with the top-100 actionability scores $a_{p|z}^{AAP}$, sample 100 actions $u_i$'s at $p$, obtain the AAP action scores $s_{u_i|z}^{AAP}$'s of these actions $u_i$'s predicted by the \textit{AAP Critic Network} $\mathcal{C}_{AAP}$, and then choose the action $u_i$ with the highest $s_{u_i|z}^{AAP}$ to execute.
We perform 10 interaction trials per test shape and report the final sample-succ rate as the percentage of sampling successful interactions in simulation.

\vspace{-3mm}
\subsection{Results and Analysis}
\vspace{-1mm}
Table~\ref{tab:numbers} presents the quantitative comparisons against the baselines showing that our method achieves the best performance in most comparison entries.
Specifically, compared to \textbf{Where2Act}, we observe that our method 
can improve the performance evidently with only 1 interaction.
Also, the performance increases as the number of interactions increases in most cases.
Compared to the \textbf{Where2Act-adaptation} baseline, 
our method with the proposed AAP module shows better performance, 
revealing that learning an adaptation network works better than using simple heuristics for adaptation.
Compared to the \textbf{Where2Act-interaction} baseline that is fed with four interactions in one shot,
our whole framework works better because our recurrent structure strategically and successively selects the most effective interaction trials.
Finally, the superior performance against the \textbf{Ours-random} and \textbf{Ours-fps} baselines that use random and FPS sampled interaction trials further validate that 
our proposed AIP module is effective in strategically and iteratively picking interaction trials.

\begin{figure*}[t]
    \begin{center}
        \includegraphics[width=\linewidth]{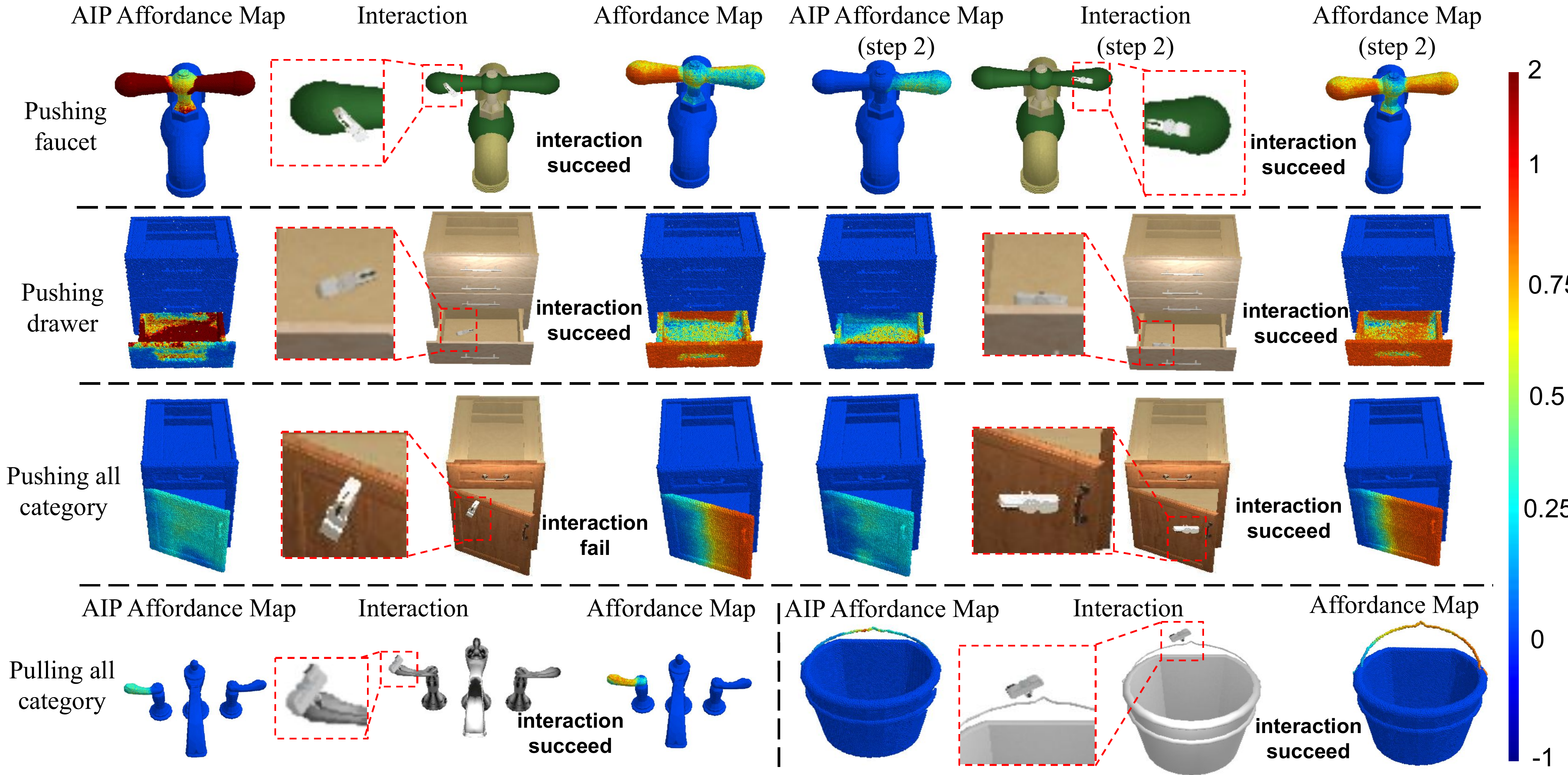}
    \end{center}
    \vspace{-3mm}
    \caption{
    Example results for the interactions proposed by the AIP module and the corresponding AIP affordance map predictions. 
    In the first three rows, we show the initial and the second AIP affordance maps, the corresponding proposed interactions, and the posterior affordance map predictions.
    In the last row, we present two more examples that only one interaction is needed.
    From these results, 
    our AIP module successfully proposes reasonable interactions for querying useful hidden information.
    }
    \vspace{-4mm}
    \label{fig:epi}
\end{figure*}

Fig.~\ref{fig:aff} shows example visualizations for our predicted affordance map posterior given interactions under different hidden kinematic or dynamic information (see the caption for more details explaining the different scenarios).
In these figures, it is clear to see that our proposed method successfully adapts the affordance prediction conditioned on different hidden information. The affordance predictions within one shape share the same visual inputs but output different results, showing that our hidden embedding $z$ contains certain information.
For example, in the first (second) column, we observe that with bigger joint friction coefficients the door (handle) is harder to manipulate and one needs to push (pull) at only points very far away from the joint axis to accomplish the task.
On the right-most (second-to-the-right) column, our network successfully figures out which side of the door (faucet switch) one needs to push.

Fig.~\ref{fig:epi} further shows some interaction proposals by our AIP module with its influence on the prediction of AAP affordance map and to the AIP affordance map itself. In the first row, for example, we see that the AIP affordance first proposes to interact at both sides of the faucet since it knows little about the hidden information but at the second timestep proposes the right side as it already learns that the left side is actionable.
Cases in the first and third rows demonstrate that the past few interactions will influence the selection of future interaction points, justifying the necessity of our recurrent structure for interaction selection. Specifically, in the third row, after the failure of the first interaction, our AIP module proposes interaction points farther from the joints since it already knows interactions on points with shorter distances than the first interaction point are not likely to succeed.  
In the last row, we show cases which only require one step to adapt.

\vspace{-2mm}
\paragraph{Ablation Study.} 
In Table~\ref{tab:ablation}, comparing against the ablated version of our method \textbf{Ours w/o iter} that trains the whole system without the interactive training process, we see that \textbf{Ours-final} achieves better results in most cases, which proves the effectiveness of the iterative training scheme. 
By iteratively alternating the training between the AAP module and the AIP module, the networks would be trained under the distribution of test-time interactions and thus achieve improved performance. 
Our method can generalize well to novel shapes and even shapes from unseen object categories through the scores in test category.

\input{tabs/ablatable}

\begin{figure*}[t]
    \begin{center}
        \includegraphics[width=\linewidth]{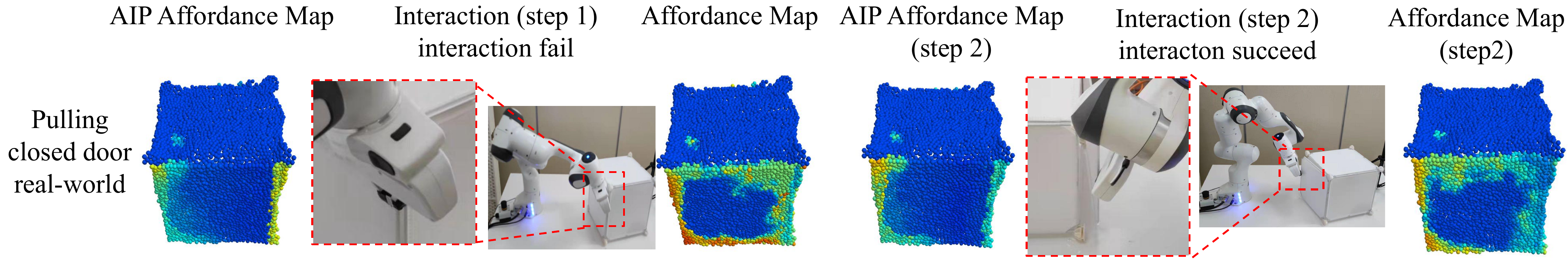}
    \end{center}
    \vspace{-3mm}
    \caption{
    Real-robot experiment on pulling open a closed door in the real world. 
    We show the AIP affordance map predictions, the AIP proposed interactions, and the AAP posterior predictions, for two interaction trials.
    The results show that our work could reasonably generalize to real-world scenarios. 
    }
    \vspace{-4mm}
    \label{fig:real}
\end{figure*}

\vspace{-2mm}
\paragraph{Real-world and Real-robot Experiments.} 
Finally, we perform real-world and real-robot experiments to show that our method can to some degree work beyond synthetic data.
We use a Franka panda robot with a two-finger parallel gripper as the actuator to pull open a cabinet door in the real world.
Fig.~\ref{fig:real} presents the results that our system proposes two interaction trials to inquire more information about this real-world cabinet and successfully learns to adapt to the posterior predictions.

Please refer to the supplementary materials for a video better illustrating this example, more experiment settings, more example results, and more experiments with additional analysis.

%% file: tabs/twotable.tex
\begin{table}[t]
  \centering
    \setlength{\tabcolsep}{3pt}
  \renewcommand\arraystretch{0.3}
  \footnotesize
   \caption{\textbf{Quantitative Evaluations.} 
  We experiment with 
  three different test-time interaction budgets (\emph{i.e.,} 1, 2, or 4) where numbers are separated by slashes.
  We use ``pushing all" and ``pulling all" to denote the experiments over all object categories, while ``pulling closed door" and ``pushing faucet" refer to the experiments over a single category only.
  For the experiments over all categories, we report the performance over novel shapes from the training categories (marked with ``train cat.") and shapes from novel categories (marked with ``test cat.").
  }
  \label{tab:numbers}
\resizebox{0.8\textwidth}{50mm}{

\begin{tabular}{@{}llccc@{}}
\toprule
 &  &  F-score (\%) & Sample-Succ (\%)\\
\midrule
\multirow{11}{*}{\shortstack[l]{pushing all\\(train cat.)}}
& Where2Act &  56.44 & 20.85 \\
& Where2Act-interaction &  72.13 & 31.53 \\
& Where2Act-adaptation &  64.16/65.42/64.99 & 20.77/22.72/26.82 \\
& ours-random & 70.24/70.58/70.85  & 29.59/31.35/32.57 \\
& ours-fps & 64.32/69.58/70.99  & 26.22/27.30/30.65 \\
& ours-final & \textbf{72.78}/\textbf{73.12}/\textbf{75.18}  & \textbf{33.82}/\textbf{33.23}/\textbf{35.23} \\
\midrule
\multirow{11}{*}{\shortstack[l]{pushing all\\(test cat.)}}
& Where2Act &  59.95 & 21.69 \\
& Where2Act-interaction &  76.12 & 37.10 \\
& Where2Act-adaptation &  51.09/53.28/55.56 & 19.06/22.27/24.50 \\
& ours-random & 75.12/76.92/76.98  & 30.78/30.78/29.48 \\
& ours-fps & 66.17/67.27/69.08  & 33.64/35.19/\textbf{37.79} \\
& ours-final & \textbf{77.58}/\textbf{77.63}/\textbf{78.42}  & \textbf{34.97}/\textbf{36.75}/37.40 \\

\midrule
\multirow{11}{*}{\shortstack[l]{pulling all\\(train cat.)}}
& Where2Act &  31.19 & 1.92 \\
& Where2Act-interaction &  38.28 & 3.89 \\
& Where2Act-adaptation &  37.22/38.48/39.13 & 1.11/2.15/1.62 \\
& ours-random & 35.03/34.48/36.84  & 4.44/2.78/6.11 \\
& ours-fps & 39.88/42.74/43.55  & 2.78/5.56/4.44 \\
& ours-final & \textbf{42.62}/\textbf{43.87}/\textbf{44.08}  & \textbf{7.78}/\textbf{9.44}/\textbf{10.55} \\

\midrule
\multirow{11}{*}{\shortstack[l]{pulling all\\(test cat.)}}
& Where2Act &  36.36 & 10.00 \\
& Where2Act-interaction &  45.80 & 9.73 \\
& Where2Act-adaptation &  40.11/45.52/48.80 & 3.40/6.25/10.17 \\
& ours-random & 41.97/44.88/46.11  & \textbf{6.13}/4.78/8.26 \\
& ours-fps & 43.67/42.77/48.33  & 4.35/3.91/4.78 \\
& ours-final & \textbf{49.51}/\textbf{50.00}/\textbf{51.33}  & 5.21/\textbf{7.39}/\textbf{10.45} \\

\midrule
\multirow{11}{*}{\shortstack[l]{pulling\\closed door}}
& Where2Act &  48.44 & 4.38 \\
& Where2Act-interaction &  66.79 & 9.09 \\
& Where2Act-adaptation &  50.21/55.75/56.81 & 6.60/7.18/6.83 \\
& ours-random & 52.41/54.25/53.37  & 7.14/6.84/6.53 \\
& ours-fps & \textbf{59.79}/63.43/69.13  & 8.88/11.33/12.10 \\
& ours-final & 57.83/\textbf{65.60}/\textbf{79.65}  & \textbf{10.86}/\textbf{11.57}/\textbf{22.14} \\

\midrule
\multirow{11}{*}{\shortstack[l]{pushing\\faucet}}
& Where2Act &  64.92 & 55.46 \\
& Where2Act-interaction &  79.85 & 80.97 \\
& Where2Act-adaptation &  66.25/62.18/67.15 & 57.50/52.08/61.70 \\
& ours-random & 72.61/76.29/79.16  & 61.81/79.01/80.82 \\
& ours-fps & 74.19/79.36/77.95  & 60.44/70.12/77.41 \\
& ours-final & \textbf{77.42}/\textbf{83.06}/\textbf{83.83}  & \textbf{65.90}/\textbf{81.66}/\textbf{82.14} \\

\bottomrule
\end{tabular}}
\vspace{-4mm}
\end{table}

%% file: tabs/ablatable.tex
\begin{table}[htb]
  \centering
    \setlength{\tabcolsep}{5pt}
  \renewcommand\arraystretch{0.3}
  \footnotesize
    \caption{\textbf{Ablation Study.} We compare our method to an ablated
version, where we remove the iteratively training process. It is clear
to see that the iteratively training process helps our framework achieve better results in most cases. 
  }

  \label{tab:ablation}
\resizebox{0.75\textwidth}{18mm}{
\begin{tabular}{@{}llccc@{}}
\toprule
 &  &  F-score (\%) & Sample-Succ (\%)\\
\midrule
\multirow{3}{*}{\shortstack[l]{pushing all (train cat.)}}
& ours w/o iter &  71.21/72.64/73.16 & 30.67/31.62/32.56 \\
& ours-final & \textbf{72.78}/\textbf{73.12}/\textbf{75.18}  & \textbf{33.82}/\textbf{33.23}/\textbf{35.23} \\
\midrule
\multirow{3}{*}{\shortstack[l]{pushing all (test cat.)}}
& ours w/o iter &  77.24/77.33/77.17 & 31.03/33.89/\textbf{38.83} \\
& ours-final & \textbf{77.58}/\textbf{77.63}/\textbf{78.42}  & \textbf{34.97}/\textbf{36.75}/37.40 \\

\midrule
\multirow{3}{*}{\shortstack[l]{pulling all (train cat.)}}
& ours w/o iter &  41.19/42.10/42.81 & 6.67/7.22/8.33 \\
& ours-final & \textbf{42.62}/\textbf{43.87}/\textbf{44.08}  & \textbf{7.78}/\textbf{9.44}/\textbf{10.55} \\

\midrule
\multirow{3}{*}{\shortstack[l]{pulling all (test cat.)}}
& ours w/o iter &  48.31/48.28/50.50 & \textbf{5.65}/6.52/9.13 \\
& ours-final & \textbf{49.51}/\textbf{50.00}/\textbf{51.33}  & 5.21/\textbf{7.39}/\textbf{10.45} \\

\midrule
\multirow{3}{*}{\shortstack[l]{pulling closed door}}
& ours w/o iter &  56.74/64.88/80.64 & 9.77/11.50/22.00 \\
& ours-final & \textbf{57.83}/\textbf{65.60}/\textbf{79.65}  & \textbf{10.86}/\textbf{11.57}/\textbf{22.14} \\

\midrule
\multirow{3}{*}{\shortstack[l]{pushing faucet}}
& ours w/o iter &  73.81/83.03/\textbf{84.32} & 61.11/81.60/\textbf{84.03} \\
& ours-final & \textbf{77.42}/\textbf{83.06}/83.83  & \textbf{65.90}/\textbf{81.66}/82.14 \\

\bottomrule
\end{tabular}}

\end{table}

%% file: tex/conclu.tex
\vspace{-3mm}
\section{Conclusion}
\vspace{-1mm}

This work addresses a big limitation of previous works learning visual actionable affordance for manipulating 3D articulated objects -- the hidden kinematic or dynamic uncertainties.
We propose a novel framework AdaAfford that samples a few test-time interactions for fastly adapting to a more accurate affordance posterior prediction removing such ambiguities.
Experimental results validate the effectiveness of our method compared to baseline approaches.

\vspace{-2mm}
\paragraph{Limitations and Future Works.}
This work only considers two action types and 3D articulated objects. Future works may study more interaction and data types.
Also, we only perform short-term interactions. Future works can investigate how to extend the framework for long-term manipulation trajectories.
Finally, we abstract away the complexity of robot arms and only use flying grippers in our experiments.
Future works shall work on considering the robot arm constraints.

%% file: tex/supp_data.tex
\section{More Detailed Data Statistics}

\begin{figure*}[htb]
    \begin{center}
        \includegraphics[width=\linewidth, 
        trim={0cm, 0cm, 0cm, 0cm}, clip]{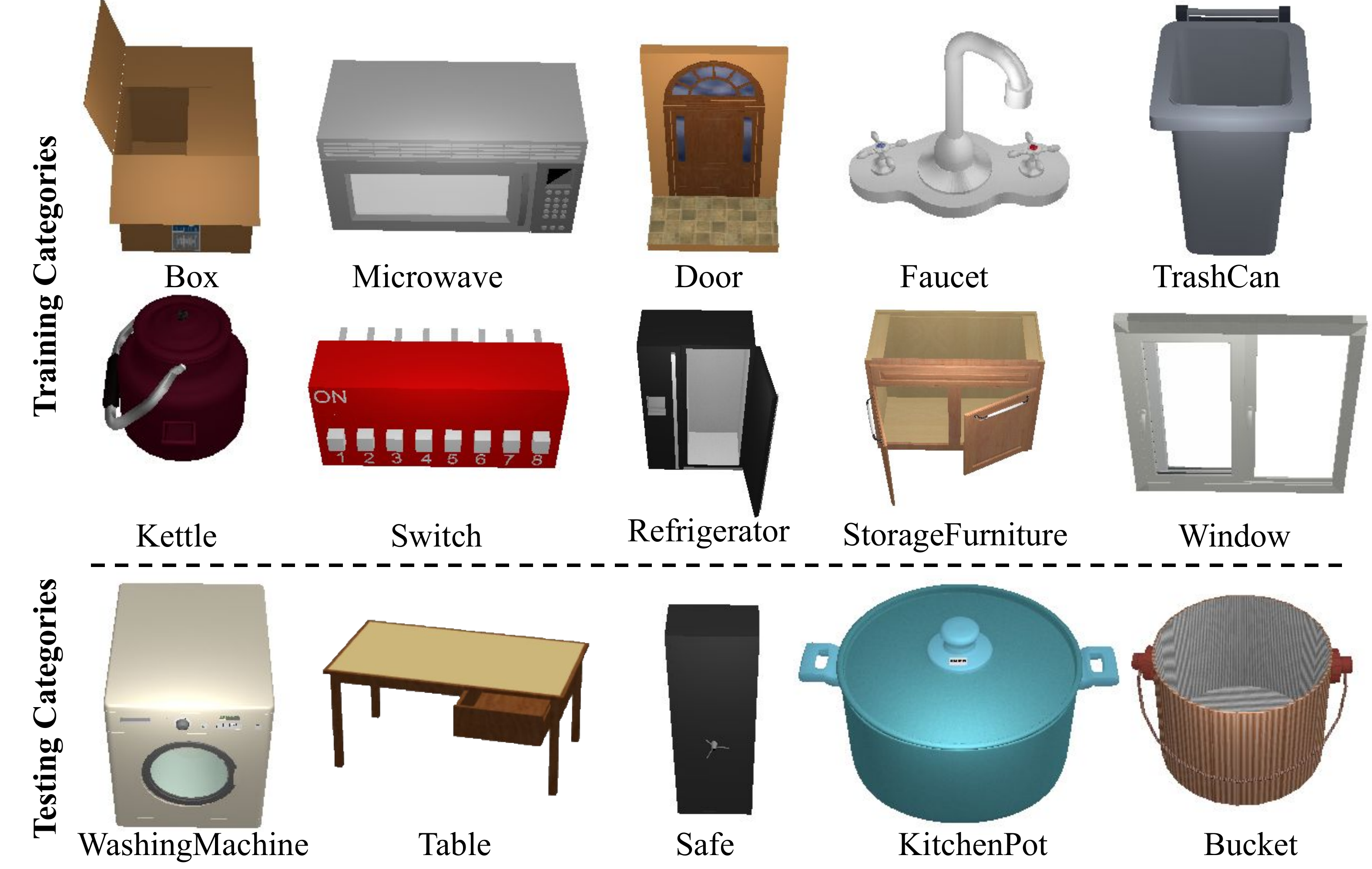}
    \end{center}
    \caption{\textbf{Data Visualization}. We show example shapes of object categories in our paper.
    }
    \label{fig:gallery}
\end{figure*}

\begin{table}[t]
  \vspace{2mm}
  \centering
  \setlength{\tabcolsep}{2pt}
  \renewcommand\arraystretch{0.8}
  \caption{We first summarize the shape counts in our dataset for pushing and pulling shapes over all categories, in which there are three data splits: training data from the training categories, test data from the training categories, and data from the test categories. We use the first split to train and use the rest two for evaluation.
  We further show the shape counts in our two additional tasks: pulling closed door and pushing faucet (denoted as Closed Door and Faucet for brevity).
  }
  \small
    \begin{tabular}{cccccc}
    \toprule
    Train-Cats  & \small{Box} & \small{Microwave} & \small{Door} & \small{Faucet} & \small{TrashCan}  \\
    \midrule
    Train / Test & 20 / 8 & 9 / 3 & 23 / 12 & 65 / 19 & 52 / 17  \\ 
    \midrule
    & \small{Kettle} & \small{Refrigerator} & \small{Switch} & \small{Cabinet} & \small{Window}  \\ 
    \midrule
    & 22 / 7 & 32 / 11 & 53 / 17 & 270 / 75 & 40 / 18  \\ 
    \midrule
    Total & 586 / 187 & & & &\\
    \toprule
    Test-Cats  & \small{Table} & \small{Washer} & \small{Bucket} & \small{Pot} & \small{Safe}\\
    \midrule
    & 95 & 16 & 36 & 23 & 29  \\
    \midrule
    Total & 199 & & & &\\
    \toprule
    ADDL Exp. & Category & Train data & Test data & &\\
    \midrule
    Closed door & Cabinet & 74 & 11 & &\\
    \midrule
    Faucet & Faucet & 15 & 4 & &\\
    \bottomrule
    \end{tabular}
  \label{tab:dataset}
\end{table}%

Table~\ref{tab:dataset} summarizes the data statistics and splits.
Figure~\ref{fig:gallery} visualizes some example shapes from our dataset.

%% file: tex/supp_ablation.tex
\section{More Experiment Settings}

\paragraph{Details of Baselines}
For \textbf{Ours-fps}, \textbf{Where2Act-interaction} and \textbf{Where2Act-adaptation} baselines, we augment the FPS method with an actionability score. In detail, we only select the points with higher actionability scores than a preset threshold (\emph{e.g.}, 0.5). If there are not enough such points, the threshold will be set lower until there exist at least 50 points whose actionability scores are higher than the threshold.
For \textbf{Where2Act-adaptation} baseline, we first train a network to give the similarities between points. For point $p_{1}$ and point $p_{2}$, given their point features extracted by PointNet++ and the distance between them, the network outputs a similarity score $sim_{p_{1}p_{2}}$. Then, an interaction $I=(O,p,R,m)$ acting on $p$ with action score $s_u$ will influence point $q$ by:
\begin{equation}
(r-s_u) * sim_{pq}
\end{equation}
where $r=1$ if $m>\tau$ (\emph{e.g.}, $\tau=0.01$) or $r=0$, $u=(p,R)$. Specifically, if the original an actionability score of point $q$ is $a_q$ and the original action score of an arbitrary action $u_*=(q,R_*)$ on point $q$ is $s_{u_*}$, the new action score $s_{u_*}^{new}$ and actionability score $a_q^{new}$ would be: 
\begin{align}
s_{u_*}^{new}=s_{u_*}+(r-s_u) * sim_{pq} \\
a_q^{new}=a_q+(r-s_u) * sim_{pq}
\end{align}
To train the network to give similarity between points, similar to our method, we use the ground truth result of the action $u_*$ as the regression target of $s_{u_*}^{new}$.

\paragraph{More baselines} We employ several baselines using FPS method to sample interaction points, and the results show the usefulness of the proposed AIP module of our framework.
\begin{itemize}
\item[-]
\textbf{Ours-purefps:} that directly uses FPS method to sample interaction points without using actionability scores.
\item[-]
\textbf{Ours-argfps:} that uses FPS augmented with actionability scores to select interaction points. When sampling a new point, we combine its distance to the sampled point set with its actionability score while doing FPS, as the weighted distance.
\end{itemize}


%% file: tex/supp_results.tex
\section{More Results and Analysis}
\begin{figure*}[t]
    \begin{center}
        \includegraphics[width=\linewidth, 
        trim={0cm, 0cm, 0cm, 0cm}, clip]{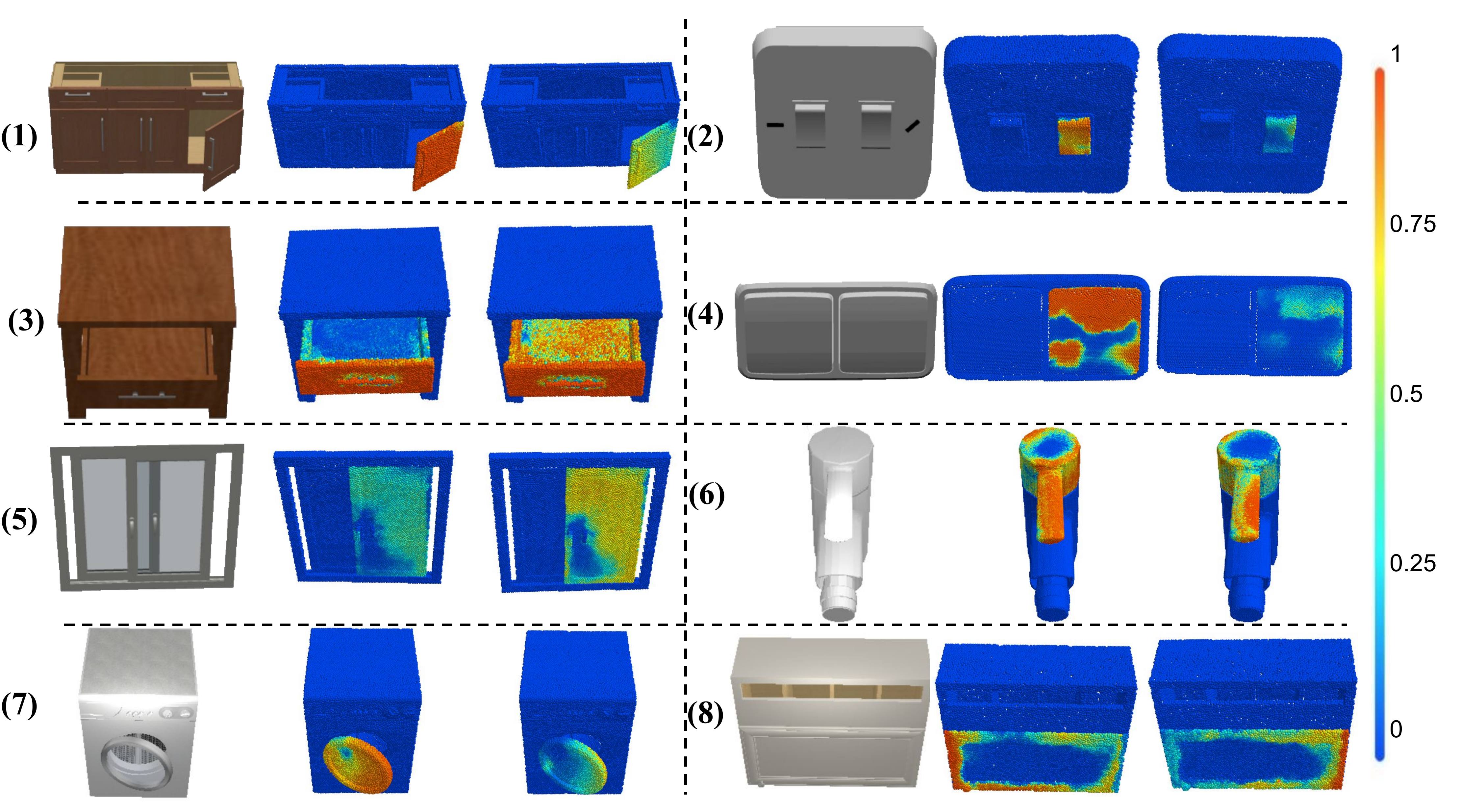}
    \end{center}
    \vspace{-3mm}
    \caption{We visualize more results for the adapted affordance predictions given by the AAP module conditioned on different hidden kinematic and dynamic information. From the first tor the last block, we respectively change the 1) mass of target part 2) joint friction 3) friction coefficient on the target part's surface 4) joint friction 5) friction coefficient on the target part's surface 6) rotating direction of the faucet 7) mass of target part 8) axis location of the door, and clearly see reasonable adaptions in affordance predictions.
    }
    \label{fig:supp_aff}
\end{figure*}

\begin{figure*}[t]
    \begin{center}
        \includegraphics[width=\linewidth, 
        trim={0cm, 0cm, 0cm, 0cm}, clip]{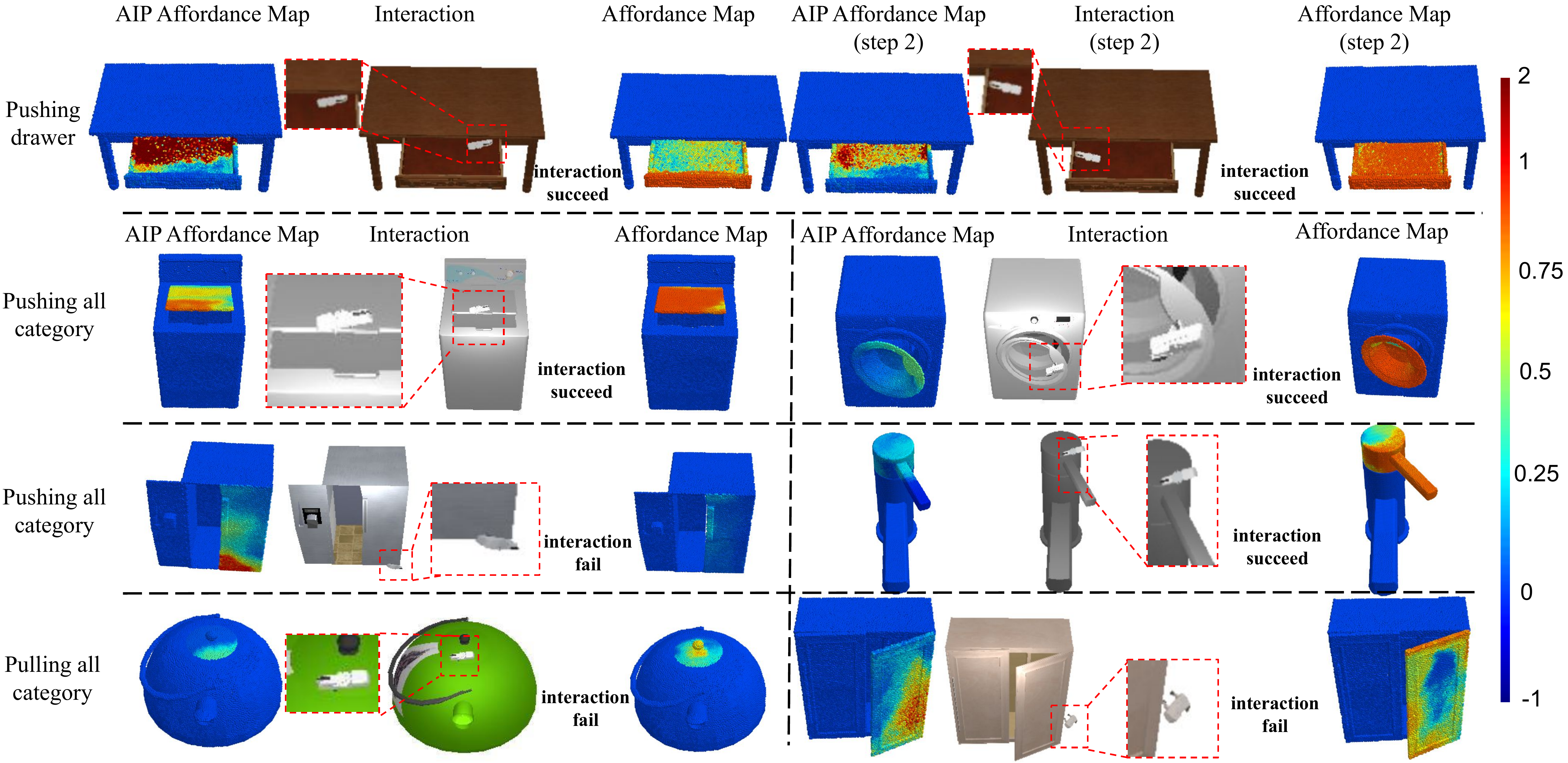}
    \end{center}
    \vspace{-3mm}
    \caption{We visualize more results for the interactions proposed by the AIP module and the corresponding AIP affordance map predictions. In the first row, we show the initial and the second AIP affordance maps, the corresponding proposed interactions, and the posterior affordance map predictions. In the last three rows, we present six more examples that only one interaction is needed. 
    }
    \label{fig:supp_ip}
\end{figure*}

In Figure~\ref{fig:supp_aff} and ~\ref{fig:supp_ip}, we show more qualitative results. See the captions of these two figures for more details.

Table~\ref{tab:fps} shows the comparisons between different methods using FPS. In most cases, both \textbf{Ours-argfps} and \textbf{Ours-fps} achieve better results than \textbf{Ours-purefps}. Because in \textbf{Ours-purefps} baseline, FPS only cares about the 3D position of points discarding the point features. While \textbf{Ours-argfps} and \textbf{Ours-fps} utilize the action scores which are generated by point features and thus achieve better results. Results show that our framework gets better performance in most cases compared with those baselines, which further shows the effectiveness of our AIP module.

\begin{table}[htb]
  \centering
    \setlength{\tabcolsep}{5pt}
  \renewcommand\arraystretch{0.3}
  \footnotesize
    \caption{\textbf{Quantitative Evaluations.}
    Comparison with different FPS baselines. Results show that our framework achieves the best performance in most cases.
  }

  \label{tab:fps}
\resizebox{0.85\textwidth}{43mm}{
\begin{tabular}{@{}llccc@{}}
\toprule
 &  &  F-score (\%) & Sample-Succ (\%)\\
\midrule
\multirow{5}{*}{\shortstack[l]{pushing all (train cat.)}}
& ours-purefps &  66.78/69.43/70.65 & 28.23/31.50/29.51 \\
& ours-argfps &  66.78/69.43/70.65 & 28.23/31.50/29.51 \\
& ours-fps &  64.32/69.58/70.99 & 26.22/27.30/30.65 \\
& ours-final & \textbf{72.78}/\textbf{73.12}/\textbf{75.18}  & \textbf{33.82}/\textbf{33.23}/\textbf{35.23} \\
\midrule
\multirow{5}{*}{\shortstack[l]{pushing all (test cat.)}}
& ours-purefps &  66.35/66.55/67.19 & 34.15/32.60/35.06 \\
& ours-argfps &  74.04/75.03/76.63 & 33.11/34.54/36.49 \\
& ours-fps &  66.17/67.27/69.08 & 33.64/35.19/\textbf{37.79} \\
& ours-final & \textbf{77.58}/\textbf{77.63}/\textbf{78.42}  & \textbf{34.97}/\textbf{36.75}/37.40 \\

\midrule
\multirow{5}{*}{\shortstack[l]{pulling all (train cat.)}}
& ours-purefps &  35.46/37.54/37.35 & 2.78/5.56/2.78 \\
& ours-argfps &  35.46/37.54/37.35 & 3.89/4.44/6.11 \\
& ours-fps &  39.88/42.74/43.55 & 2.78/5.56/4.44 \\
& ours-final & \textbf{42.62}/\textbf{43.87}/\textbf{44.08}  & \textbf{7.78}/\textbf{9.44}/\textbf{10.55} \\

\midrule
\multirow{5}{*}{\shortstack[l]{pulling all (test cat.)}}
& ours-purefps &  43.60/48.91/47.36 & 6.96/5.22/3.91 \\
& ours-argfps &  45.17/47.39/50.60 & \textbf{8.69}/7.22/10.00 \\
& ours-fps &  43.67/42.77/48.33 & 4.35/3.91/4.78 \\
& ours-final & \textbf{49.51}/\textbf{50.00}/\textbf{51.33}  & 5.21/\textbf{7.39}/\textbf{10.45} \\

\midrule
\multirow{5}{*}{\shortstack[l]{pulling closed door}}
& ours-purefps &  53.53/59.81/67.20 & 6.67/7.64/10.71 \\
& ours-argfps &  58.42/62.31/68.72 & 8.94/11.25/13.75 \\
& ours-fps &  \textbf{59.79}/63.43/69.13 & 8.88/11.33/12.10 \\
& ours-final & 57.83/\textbf{65.60}/\textbf{79.65}  & \textbf{10.86}/\textbf{11.57}/\textbf{22.14} \\

\midrule
\multirow{5}{*}{\shortstack[l]{pushing faucet}}
& ours-purefps &  73.39/79.13/79.85 & 61.88/76.59/72.50 \\
& ours-argfps &  74.66/78.30/79.61 & 61.42/66.65/74.75 \\
& ours-fps &  74.19/79.36/77.95 & 60.44/70.12/77.41 \\
& ours-final & \textbf{77.42}/\textbf{83.06}/\textbf{83.83} & \textbf{65.90}/\textbf{81.66}/\textbf{82.14} \\

\bottomrule
\end{tabular}}

\end{table}

